\documentclass[10pt,twocolumn,letterpaper]{article}
\usepackage[table,xcdraw]{xcolor}
\usepackage{wacv}
\usepackage{times}
\usepackage{epsfig}
\usepackage{graphicx}
\usepackage{amsmath}
\usepackage{amssymb}
\usepackage{flafter}
\usepackage{caption}
\usepackage{float}
\usepackage{siunitx}
\usepackage{algorithm,algorithmic}
\usepackage{booktabs}
\usepackage{threeparttable}
\usepackage{multirow}
\usepackage{makecell}
\usepackage{placeins}
\usepackage{mathrsfs} 
\usepackage{fixltx2e}
\usepackage{color}
\usepackage{caption}
\usepackage{bm}
\usepackage{wrapfig}
\usepackage{setspace,lipsum}
\usepackage{marginnote}
\usepackage{hhline}
\usepackage{multirow}

\newcommand{\R}{\mathbb{R}}
\newcommand{\N}{\mathcal{N}}
\newcommand{\Sp}{\mathcal{S}}

\newcommand\norm[1]{\left\lVert#1\right\rVert}
\DeclareSIUnit\Molar{\textsc{m}} 
\DeclareSIUnit{\pH}{pH}
\DeclareSIUnit{\pixel}{px}

\renewcommand{\vec}[1]{\text{\boldmath$#1$}}

\DeclareMathOperator*{\argminA}{arg\,min} 
\DeclareMathOperator*{\argmaxA}{arg\,max} 

\DeclareMathOperator*{\argmax}{\arg\!\max}

\wacvfinalcopy 

\def\wacvPaperID{680} 

\ifwacvfinal\fi
\begin{document}

\title{Rethinking Monocular Depth Estimation with Adversarial Training}

\author{Richard Chen$^1$, Faisal Mahmood$^2$, Alan Yuille$^1$ and Nicholas J. Durr$^2$\\
$^1$Department of Computer Science $^2$Department of Biomedical Engineering \\
Johns Hopkins University, Baltimore, MD\\
{\tt\small \{rchen40, faisalm, ayuille, ndurr\}@jhu.edu}}

\maketitle
\ifwacvfinal\thispagestyle{empty}\fi

\begin{abstract}
   Monocular depth estimation is an extensively studied computer vision problem with a vast variety of applications. Deep learning-based methods have demonstrated promise for both supervised and unsupervised depth estimation from monocular images. Most existing approaches treat depth estimation as a regression problem with a local pixel-wise loss function. In this work, we innovate beyond existing approaches by using adversarial training to learn a context-aware, non-local loss function. Such an approach penalizes the joint configuration of predicted depth values at the patch-level instead of the pixel-level, which allows networks to incorporate more global information. In this framework, the generator learns a mapping between RGB images and its corresponding depth map, while the discriminator learns to distinguish depth map and RGB pairs from ground truth. This conditional GAN depth estimation framework is stabilized using spectral normalization to prevent mode collapse when learning from diverse datasets. We test this approach using a diverse set of generators that include U-Net and joint CNN-CRF. We benchmark this approach on the NYUv2, Make3D and KITTI datasets, and observe that adversarial training reduces relative error by several fold, achieving state-of-the-art performance. 
\end{abstract}


\section{Introduction}

\begin{figure}
\centering
\includegraphics[width=3.2in]{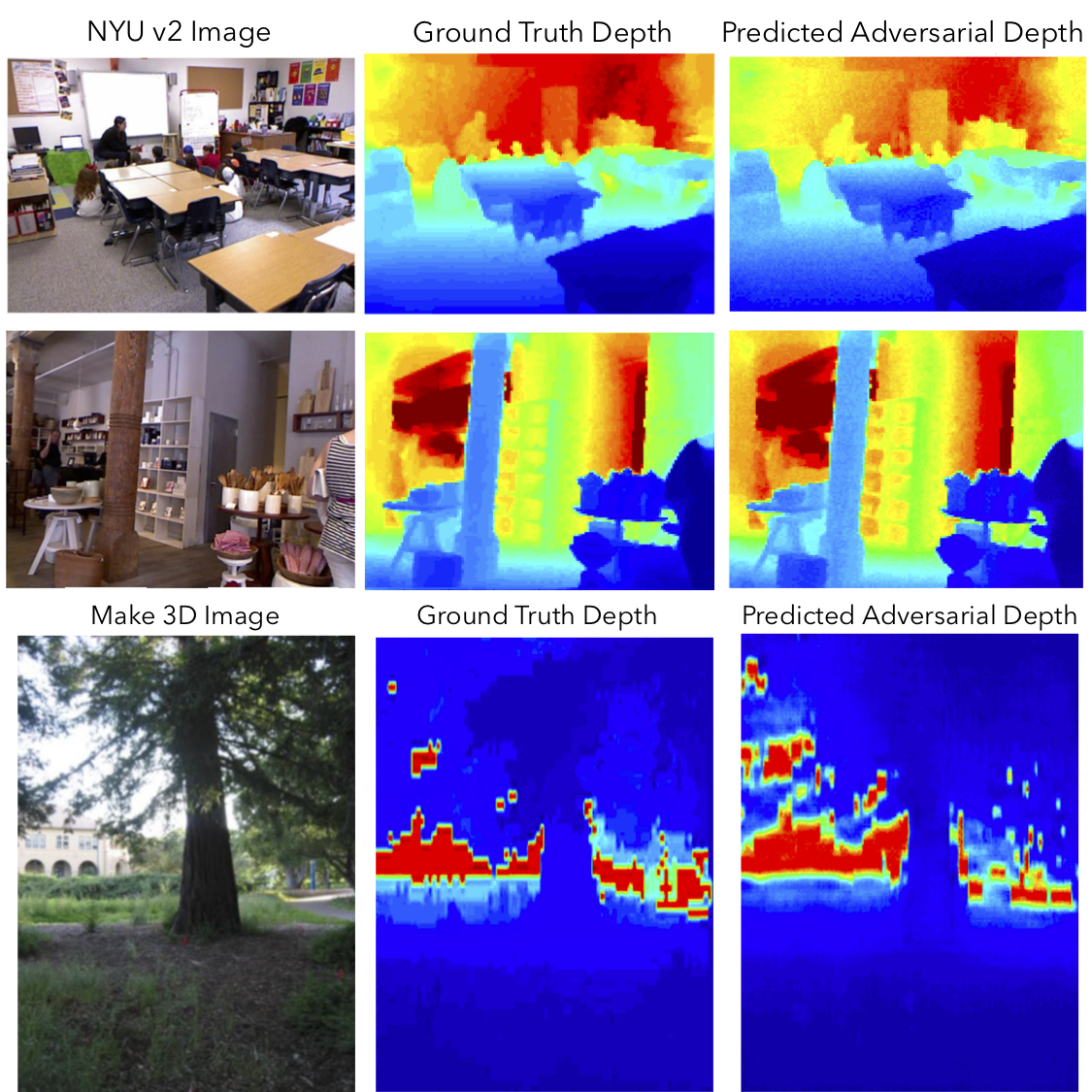}
\caption{Representative images showing estimated depth on NYUv2 and Make 3D datasets via our proposed adversarial depth estimation paradigm.}
\label{fig_sim}
\end{figure}

Depth estimation is one of the most extensively studied tasks by the computer vision community, largely due to its value in facilitating scene understanding and geometric relations between objects \cite{li2015depth,liu2010single,eigen2015predicting}. Fusing depth has demonstrated improved performance on a number of computer vision tasks including semantic segmentation, topographical reconstruction, and activity recognition \cite{wang2015towards,luvizon20182d,hazirbas2016fusenet}. Previously, the computer vision community relied on multiview methods such as stereo vision and structure-from-motion for depth estimation \cite{okutomi1993, barnard1982, Baker1981, nayar1990shape, weinshall1988application, aloimonos1988shape}. However, situations where multiple measurements from the same scene may not be available or difficult to acquire motivate the need for developing monocular depth estimation methods.

Though deep networks have shown promise in estimating depth from monocular images, many methods rely on local pixel-wise loss functions that do not capture higher-order statistics of the training data. To make these networks more context-aware, many loss functions calculate image gradients to capture changes in depth and preserve structural details \cite{eigen2015predicting,saxena2007learning,hoiem2007recovering}. This problem has also been partially addressed by the many combinations of deep learning and graphical model-based methods \cite{liu2016learning,fu2018deep,mahmood2017unsupervised,mahmood2017deep}. CNN-graphical model setups such as jointly trained CNN-CRF methods are more context-aware as compared to regular CNNs. While hybrid CNN-CRF models such as Liu \textit{et al.} \cite{liu2016learning} and Mahmood \textit{et al.} \cite{mahmood2017unsupervised,mahmood2017deep} maintain some spatial consistency between the prediction and the ground truth depth map via the pairwise potential, over-segmenting the image into super-pixels prevents the network from learning higher-order statistics that may describe depth cues in the image.

Recently, conditional generative adversarial networks (cGANs) have become an emerging technique in learning mapping distributions of high-dimensional data \cite{isola2017image}. Such methods have mainly been used for image-to-image translation tasks such as artistic style transfer, super-resolution \cite{bulat2017super}, and synthetic data refinement \cite{shrivastava2017learning}. However, they can also be used in inference tasks such as semantic segmentation\cite{luc2016semantic}, in which the generator learns a mapping from objects to their semantic labels in an image, and the discriminator provides feedback to the generator about its accuracy. As argued by Luc \textit{et al.} and Isola \textit{et al.} \cite{luc2016semantic, isola2017image}, this adversarial term can be interpreted as a non-local loss that penalizes the joint configuration of pixel values. We argue further that this non-local loss, when calculated at the patch-level, is beneficial for learning depth cues. From our experiments, the benefits of conditional adversarial learning are two fold: a) networks can learn a loss function for depth estimation, which promotes the recovery of features that would be generally lost due to the limitations of a local pixel-wise loss function, and (b) such a setup is more context-aware, as the discriminator forces the generator to generate realistic pixel configurations of predicted depth that would be indistinguishable from ground truth depth maps.


\textbf{Contributions:} In this work, we propose that deep learning-enabled monocular depth estimation can be enhanced with adversarial training. We demonstrate an improvement over state-of-the-art depth estimation results on NYUv2, Make3D and KITTI datasets using conditional GANs, and investigate how the addition of an adversarial term affects performance when using encoder-decoder network and CNN-graphical model setups as generators. The specific contributions of our work are summarized below: 

\begin{enumerate}
 \item  We describe a framework for context-aware depth estimation with stable adversarial training. Our framework learns a non-local loss function for depth estimation by incorporating a patch-level adversarial term, in which the discriminator classifies regions in the depth map predictions as synthetic or realistic. Depth regions that are predicted as synthetic penalize the generator for producing unrealistic depth configurations that fail to mimic real depth regions.
 \item  We present state-of-the-art algorithms and models for monocular depth estimation on the NYUv2, Make3D and KITTI datasets, and demonstrate how adversarial training can be adapted for different model types.
\end{enumerate}

\section{Related Work}

\noindent\textbf{Monocular Depth Estimation.}\\
Historically, depth estimation has been approached by multiview methods such as stereopsis \cite{weinshall1988application}. A large body of knowledge has also focused on recovering depth from shading \cite{prados2006shape}, texture \cite{aloimonos1988shape}, and focus \cite{nayar1990shape}. Many approaches for monocular depth estimation rely on hand-crafted features, probabilistic graphical models, and deep networks to extract multi-scale contextual information in scenes. Prior to deep networks, depth estimation became posed as a Markov Random Field (MRF) learning problem. Saxena \textit{et al.} \cite{saxena2007learning} used a patch-based MRF to model relations between the depth of image patches with its immediate neighbors at different scales. Liu \textit{et al.} \cite{liu2010single} used both semantic and superpixel segmentation information from single images to help guide depth perception, using a pixel-based MRF and superpixel-based MRF to incorporate semantic and geometric constraints respectively. Ladicky \textit{et al.} \cite{ladicky2014pulling} also incorporated semantic information by learning a joint classifier to predict both depth and semantic labels.

Following the breakthrough performance of CNNs for regression and classification tasks, depth estimation is often posed as a regression problem using end-to-end trained deep networks, with some recent efforts being made combine deep networks with graphical models. Eigen \textit{et al.} \cite{eigen2015predicting} was the first to use CNNs for monocular depth estimation, in which they proposed a multi-scale deep network that first generates a coarse depth using a fully connected layer, followed by a refinement network that recovers texture details. Laina \textit{et al.} \cite{laina2016deeper} adopted a fully convolutional architecture that learns an upsampling convolution layer instead of a fully-connected layer to obtain finer depth estimates at higher resolutions, and exploits network depth to capture global information in an image.  Liu \textit{et al.} \cite{liu2016learning} presented a CNN-CRF network where the unary potential is a regression term that predicts depth for a given superpixel using fully convolutional layers, and the pairwise potential is a smoothness term measures intensity, color and texture differences between neighboring superpixels. Wang \textit{et al.} \cite{wang2015towards} introduced a hierarchical CNN-CRF that jointly predicts depth and semantic segmentation from the same features, and was able to refine superpixel-wise CNN depth predictions. Xu \textit{et al.} \cite{xu2018monocular} learned multi-scale representations by recovering depth maps at each side output of an encoder-decoder network using a continuous CRF framework, and later followed their work by incorporating attention modules at the bottleneck of their encoder-decoder network \cite{xu2018structured}.

In addition to advancements made in neural network architectures to incorporate context, there is also interest in engineering novel loss functions for recovering depth beyond $\mathcal{L}_{\text{L1}}$, $\mathcal{L}_{\text{L2}}$, and Huber (Smooth $\mathcal{L}_{\text{L1}}$). Eigen \textit{et al.} \cite{eigen2015predicting} was the first to use $\mathcal{L}_{\text{L1}}$ loss in \textit{log}-space and gradient loss terms in deep networks. The use of \textit{log}$\mathcal{L}_{\text{L1}}$ downweights the contribution of depth errors at background pixel indices which tend to have less rich information, and the use of gradient terms help preserve details on local structure and surface regions. Laina \textit{et al.} \cite{laina2016deeper} introduced the BerHu loss, which penalized low and high errors by an $\ell_1$-norm and $\ell_2$-norm respectively. Jiao \text{et al.} \cite{Jiao2018ECCV} observed that on some depth estimation benchmarks, the distribution of depth values are skewed towards the foreground, which motivated a different loss function than Eigen \textit{et al.} that weighted depth errors at the background indices more heavily than those at the foreground indices.\\

\noindent\textbf{Conditional Generative Adversarial Networks.}
\par
The GAN framework was first presented by Goodfellow \textit{et. al.} in \cite{salimans2016improved,goodfellow2016nips,goodfellow2014generative} and was based on the idea of training two networks, a generator and a discriminator simultaneously with competing losses. While the generator learns to generate realistic data from a random vector, the discriminator classifies the generated image as real or fake and gives feedback to the generator. GANs have recently been used for a variety of different applications \cite{sbai2018design,choi2018real,burlingame2018shift,ahsan2018discrimnet} including image-to-image translation \cite{isola2017image} and style-transfer and synthetic data generation \cite{mahmood2017unsupervised}. Although, GANs have a generative and artistic ability, in order to harness the benefits of the GAN framework for specific vision applications they must be conditioned by additional information. This auxiliary information can be class labels, images or any other kind of data. Such a setup is termed conditional GANs (cGANs) and was first introduced by Mirza \textit{et al.}  \cite{mirza2014conditional}. In cGANs, the noise vector typical to GAN problems is combined with this auxiliary conditioning information resulting in generative models that are capable of transferring between domains. This approach has been used for paired \cite{isola2017image} and unpaired image-to-image translation \cite{zhu2017unpaired}. Since their advent, cGANs have been used for a variety of computer vision tasks, most notably in semantic segmentation \cite{luc2016semantic}. Recently, Krishna \textit{et al.}  \cite{regmi2018cross} used cGANs for cross view image synthesis. Wang \textit{et al.} \cite{wang2018stacked} have used cGANs for jointly learning shadow detection and removal and Hong \textit{et al.} \cite{hong2018conditional} used it for structured domain adaptation. In the joint landscape of both monocular depth estimation and GANs, Pilzer \textit{et al.} \cite{Pilzer2018} describes an unsupervised approach for depth estimation using cycle-consistent adversarial training. Because this approach uses unpaired data, the adversarial training does not explicitly preserve structural information and surface regions from depth prediction and its ground truth \cite{Hartley00}.

\begin{figure}[h]
\centering
\includegraphics[width=3.8in]{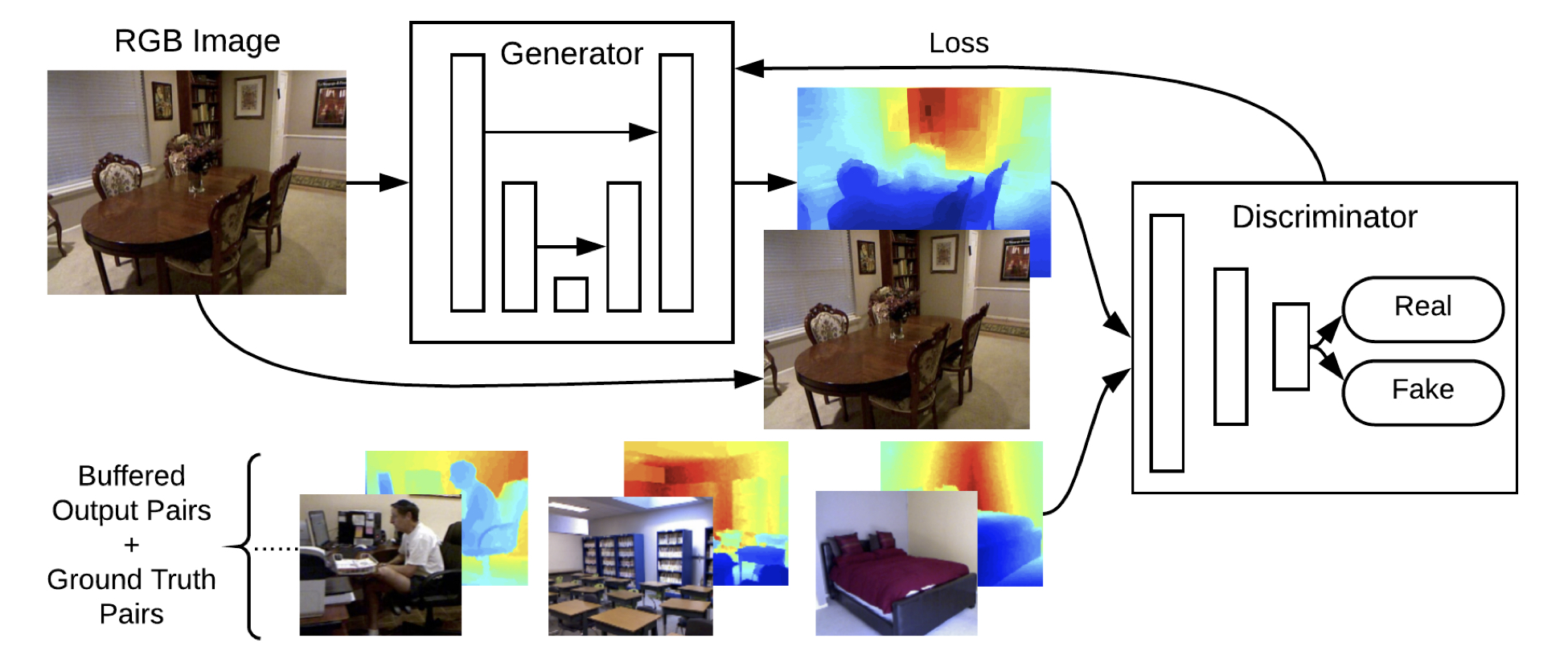}
\caption{Conditional GAN-based depth estimation architecture with U-Net as a generator.}
\label{fig_sim}
\end{figure}

\section{Conditional GAN Framework for Depth\\ Estimation}

In this section, we describe the conditional GAN objective for training depth estimation networks with non-local adversarial loss, followed by network architecture details. We denote $A$ and $A_d$ as the RGB and depth image domains respectively, and $a$ and $a_d$ as training examples in $A$ and $A_d$. Additionally, we denote $G$ as a mapping function $G: A\rightarrow A_d$ that learns a mapping from RGB to depth, and $D$ as the discriminator network for $G$.

\subsubsection{Conditional GAN Objective}

The conditional GAN framework consists of two networks that compete against each other in a \textit{min-max} game to respectively minimize and maximize the objective, $\text{min}_G \text{max}_{D} \mathcal{L}(G, D)$. The generator $G$ learns a mapping from $A$ to $A_d$, and the discriminator $D$ distinguishes between real and synthesized pairs of depth and RGB. To train this framework for depth estimation for paired data, the conditional GAN objective consists of an adversarial loss term $\mathcal{L}_{\text{GAN}}$ and a per-pixel loss term  $\mathcal{L}_\text{L1}$ to penalize both the joint configuration of pixels and accuracy of the estimated depth maps.

The adversarial loss is used to match the distribution of generated samples to that of the target distribution. For the mapping $G: A \rightarrow A_d$, we can express the adversarial objective as the binary cross entropy loss of $D$ in classifying real/synthesized pairs. We can express this loss as:

\begin{equation}\label{eq:Tk}
  \begin{aligned}
\text{min}_G \text{max}_{D} \mathcal{L}_{\text{GAN}}(G, D) = \mathbb{E}_{a,a_d\sim p_{\text{data}}(a,a_d)
}[\text{log}D(A,A_d)] +\\ \mathbb{E}_{a\sim p_{\text{data}}(a,g(a))}[\text{log}(1-D(A,G(A))]
  \end{aligned}
\end{equation}

The $L_1$ loss term is used to score the accuracy of the depth estimation by $G$,

\begin{equation}\label{eq:Tk}
  \begin{aligned}
 \mathcal{L}_{\text{L1}}(G) = \mathbb{E}_{a,a_d\sim p_{\text{data}}}(a,a_d)[||a_d-G(a)||_1] 
  \end{aligned}
\end{equation}

The motivation for using an adversarial loss is to incorporate non-local information, which has been shown to be instrumental for monocular depth estimation. We train a generator to learn a mapping between a RGB image and its corresponding depth map, and a discriminator to distinguish between ground truth and predicted depth conditioned on the RGB image on the patch-level.  Per-pixel losses are generally local, in that each output pixel is considered conditionally independent from all other pixels given the image. When used in depth estimation, per-pixel losses such as $\mathcal{L}_{\text{L1}}$ and $\mathcal{L}_{\text{L2}}$ tend to produce blurry results, as the total relative error is averaged across all pixels which. An adversarial loss, on the other hand, penalizes the joint configuration of pixel predictions made in an image. The adversarial loss can be interpreted as a non-local loss that can help preserve more details. The adversarial loss comes from the discriminator, which classifies overlapping pairs of image and depth patches as being real or synthetic. By controlling the size of the patch, we can control the size of the non-locality, with bigger patches incorporating more global information in the image. Experimentally, we observed that predicting on $70 \times 70$-sized patches allowed the generator to make fine-grained depth predictions. Thus, we can write the loss function for the conditional GAN framework as:

\begin{equation}\label{eq:Tk}
  \begin{aligned}
    \argminA_G \argmaxA_D \mathcal{L}_{\text{GAN}}(G,D) + \lambda\mathcal{L}_{\text{L1}}(G)
  \end{aligned}
\end{equation}

\noindent where, $\lambda$ is a mixing parameter. As argued in Isola \textit{et al.} \cite{isola2017image}, such an adversarial loss setup can be thought as a learned loss function where the adversarial loss is learned as the discriminator and generator are trained. 

\subsection{Stabilizing GAN training}
As noted in previous works, the training procedure for GANs can be very unstable and lead to mode collapse and gradient artifacts in the depth predictions\cite{goodfellow2016nips}. To address this, we apply spectral normalization in both the generator and the discriminator. Spectral normalization was first introduced in Miyamoto et al. \cite{miyato2018spectral}, and was used to control the Lipschitz constant of the discriminator such that the spectral norm $\sigma$ of the convolution weights $W$ in the network would be bounded by the Lipschitz constraint: $\sigma(W) = 1$. As a result, the discriminator is more stable during training and can avoid exploding gradients. Following the two-timescale update rule in Heusal \textit{et al.} \cite{heusel2017gans}, the learning rate for the discriminator was set to be four times the learning rate of the generator to increase speed of convergence.  In subsequent experiments by \cite{zhang2018self}, spectral normalization was empirically determined to be also beneficial for stabilizing the generator, allowing for fewer discriminator updates per generator update. We also further stabilize the discriminator by using a buffered data input from the generator, which consists of previously generated and classified pairs and ground truth data. This approach to stabilizing the GAN training procedure was presented in Shrivastava et al.\cite{shrivastava2017learning}, and has been used in several proceeding works \cite{isola2017image, mahmood2017unsupervised}. In our observations, we found that these techniques reduced visual artifacts made by the generator, which resulted in more smoothly-varying depth estimates.

\begin{figure}[h]
\centering
\includegraphics[width=3.4in]{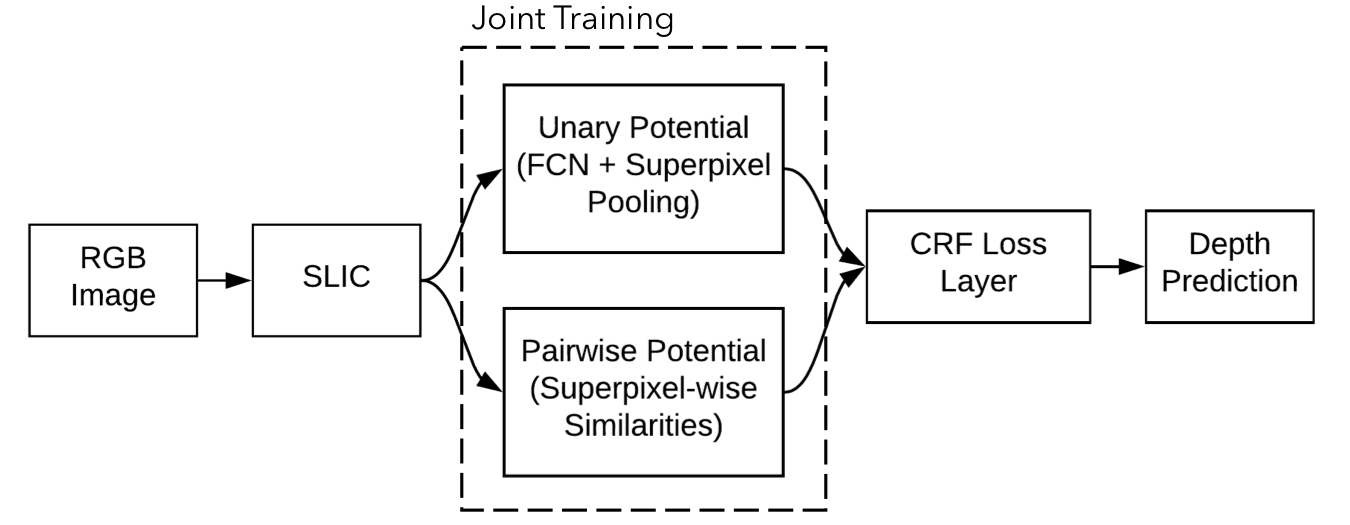}
\caption{Joint CNN-CRF training paradigm. CNN-CRF models are more context aware as compared to regular CNN models. However, due to computational complexity we used this model as a generator on a super-pixel level.}
\label{fig_sim}
\end{figure}

\begin{table*}[!h]
\begin{center}
\begin{tabular*}{\linewidth}{@{\extracolsep{\fill}} c||c c c|c c c}
\hline
Method & rel $\downarrow$& $log_{10} $ $\downarrow$& rms $\downarrow$& $\delta < 1.25$ $\uparrow$& $\delta < 1.25^2$$\uparrow$ & $\delta < 1.25^3$$\uparrow$ \\
\hline\hline
Make3D\cite{saxena2007learning}        & 0.349 & - & 1.214  & 0.447 & 0.745 & 0.897      \\
DepthTransfer\cite{karsch2016depth}       & 0.350 & 0.131 & 1.20     & -    & - & - \\
Liu \textit{et al.} \cite{liu2010single}        & 0.335 & 0.127 & 1.06 & - & - & -\\
Ladicky \textit{et al.} \cite{ladicky2014pulling}              & - & - & -     & 0.542  & 0.829 & 0.941 \\
Li \textit{et al.} \cite{li2015depth}    & 0.232 & 0.094 & 0.821  & 0.621 & 0.886 & 0.968      \\
Wang \textit{et al.} \cite{wang2015towards} & 0.220 & - & 0.824     & 0.605 & 0.890 & 0.970       \\
Roy \textit{et al.} \cite{roy2016monocular}         & 0.187 & - & 0.744    & - & - & -           \\
Liu \textit{et al.} \cite{liu2016learning}     & 0.213 & 0.087 & 0.759   & 0.650 & 0.906 &  0.976      \\
Eigen \textit{et al.} \cite{eigen2015predicting}        & 0.158 & - & 0.641 & 0.769 & 0.950 & 0.988           \\
Chakrabarti \textit{et al.} \cite{chakrabarti2012depth}     & 0.149 & - & 0.620   & 0.806 & 0.958 & 0.987           \\
Laina \textit{et al.} \cite{laina2016deeper}         & 0.194 & 0.083 & 0.790     & 0.629 & 0.889 & 0.971      \\
Li \textit{et al.} \cite{li2017two}           & 0.152 & 0.064 & 0.611   & 0.789 & 0.955 & 0.988        \\
MS-CRF \textit{et al.} \cite{xu2018monocular}         & 0.121 & 0.052 & 0.586    & 0.811 & 0.954 & 0.987      \\
DORN \cite{fu2018deep}                    & 0.115 & 0.051 & 0.509   & 0.828 & 0.965 & 0.992           \\
\hline
CNN-CRF      					 	 & 0.232	& 0.094 & 0.824    & 0.614 & 0.883 & 0.971    \\
\textbf{CNN-CRF-Adv.}         	 & 0.202	& 0.081 & 0.755  & 0.658 & 0.901 & 0.962  \\
U-Net \cite{ronneberger2015u}    & 0.327	& 0.124 & 0.981    & 0.508 & 0.783 & 0.815    \\
\textbf{U-Net-Adv.} 			 & \textbf{0.114}	& \textbf{0.050} & \textbf{0.4871}    & \textbf{0.852} & \textbf{0.971} & \textbf{0.997}\\
\hline
\end{tabular*}
\end{center}
\caption{Performance on NYUv2. All methods are evaluated on the test split by Eigen \textit{et al.} \cite{eigen2015predicting}}
\end{table*}

\subsection{Network Architectures}
In this section, we provide details of the two different types of network architectures used as a generator for depth estimation with adversarial training.\\

\noindent\textbf{Encoder-Decoder Networks.} Encoder-decoder networks  are commonly used in many deep network approaches for monocular depth estimation \cite{laina2016deeper,xu2018monocular,xu2018structured,Jiao2018ECCV,odena2016}. One formulation, the U-Net achitecture by Ronneberger \textit{et al.} \cite{ronneberger2015u}, draws skip connections between convolution layers on the encoder path and up-sampling layers on the decoder path that have the same spatial size. These connections are made between feature maps to recover and enforce spatial information across multiple resolutions and enforce spatial consistency on the output image, where the input and outputs are expected to align channel-wise  \cite{drodzdal2016}. We demonstrate that introducing an adversarial term can recover higher order information, while also preserving object boundaries and shape details (Fig. 4, Table 1). Our implementation of U-Net (Fig. 2) assumes that input images are $256 \times 256$, as the inputs are down sampled to $1 \times 1$ pixel at the bottleneck. Pooling and up-sampling operations are replaced with $4 \times 4$ convolution filters with stride $2 \times 2$ and transposed convolutions respectively. The U-Net loss can be defined as,
\begin{equation}\label{eq:Tk}
  \begin{aligned}
\argminA_{G=\text{U-Net}} \argmaxA_D \mathcal{L}_{\text{GAN}}(G_{\text{U-Net}},D) + \lambda\mathcal{L}_{\text{L1}} (G_{\text{U-Net}}).
  \end{aligned}
\end{equation}

\noindent\textbf{Joint CNN-CRF Network.}
In this section we explain the how an adversarial loss can be used in a joint CNN-CRF network. Assuming $\vec{x}\in \R^{n\times m}$ be an image which has been divided into $g$ superpixels and $\vec{y}=[y_1,y_2,...,y_g] \in \R$ be the depth vector corresponding each superpixel. In this case the conditional probability distribution of the raw data can be defined as, 

\begin{equation}\label{eq:Tk}
  \begin{aligned}
 {Pr(\vec{y}|\vec{x})}=\frac{exp(E(\vec{y},\vec{x}))}{\int_{-\infty}^{\infty} exp(E(\vec{y},\vec{x})) d\vec{y}}.
  \end{aligned}
\end{equation}

 \noindent and $E$ is the energy function. In order to predict the depth of a new image we must solve the maximum aposteriori (MAP) problem, $\widehat{\vec{y}}=\argmax_{y} {Pr(\vec{y}|\vec{x})}.$

Let $\psi$ and $\phi$ be unary and pairwise potentials over superpixel nodes $\N$ and edges $\Sp$ of $\vec{x}$, then the energy function can be formulated as,

\begin{equation}\label{eq:Tk}
  \begin{aligned}
 E(\vec{y},\vec{x}) = \sum_{i \in \N} \psi({y}_i,\vec{x};\vec\gamma) + \sum_{(i,j) \in \Sp} \phi({y}_{i},{y}_{j},\vec{x};\vec\beta),
  \end{aligned}
\end{equation}

 \noindent where, $\psi$ regresses the depth from a single superpixel and $\phi$ encourages smoothness between neighboring superpixels. The objective is to learn the two potentials in a unified convolutional neural network (CNN) framework. This setup is shown in Fig. 3. The unary part takes a single image superpixel patch as an input and feeds it to a CNN which outputs the regressed depth of that superpixel. Based on \cite{liu2016learning,mahmood2017deep} the unary potential can be defined as,

\begin{equation}\label{eq:Tk}
  \begin{aligned}
 \psi(y_{i},\vec{x};\vec\gamma)= - (y_i-h_i(\vec{\gamma}))^2
  \end{aligned}
\end{equation}

\noindent where $h_i$ is the regressed depth of superpixel $i\in N$ and $\gamma$ represents CNN parameters. The pairwise potential function is based on the standard CRF vertex and edge feature function studied in \cite{mahmood2017unsupervised}. Let $\vec{\beta}$ be the network parameters and $\vec{S}$ be the similarity matrix where ${S}_{i,j}^k$ represents $k$ similarity metrics between the $i^{th}$ and $j^{th}$ superpixel. We use the intensity difference and grayscale histogram as pairwise similarity metrics using $\ell_2$-norm. The pairwise potential can be defined as,

\begin{figure*}
\centering
\includegraphics[width=\textwidth]{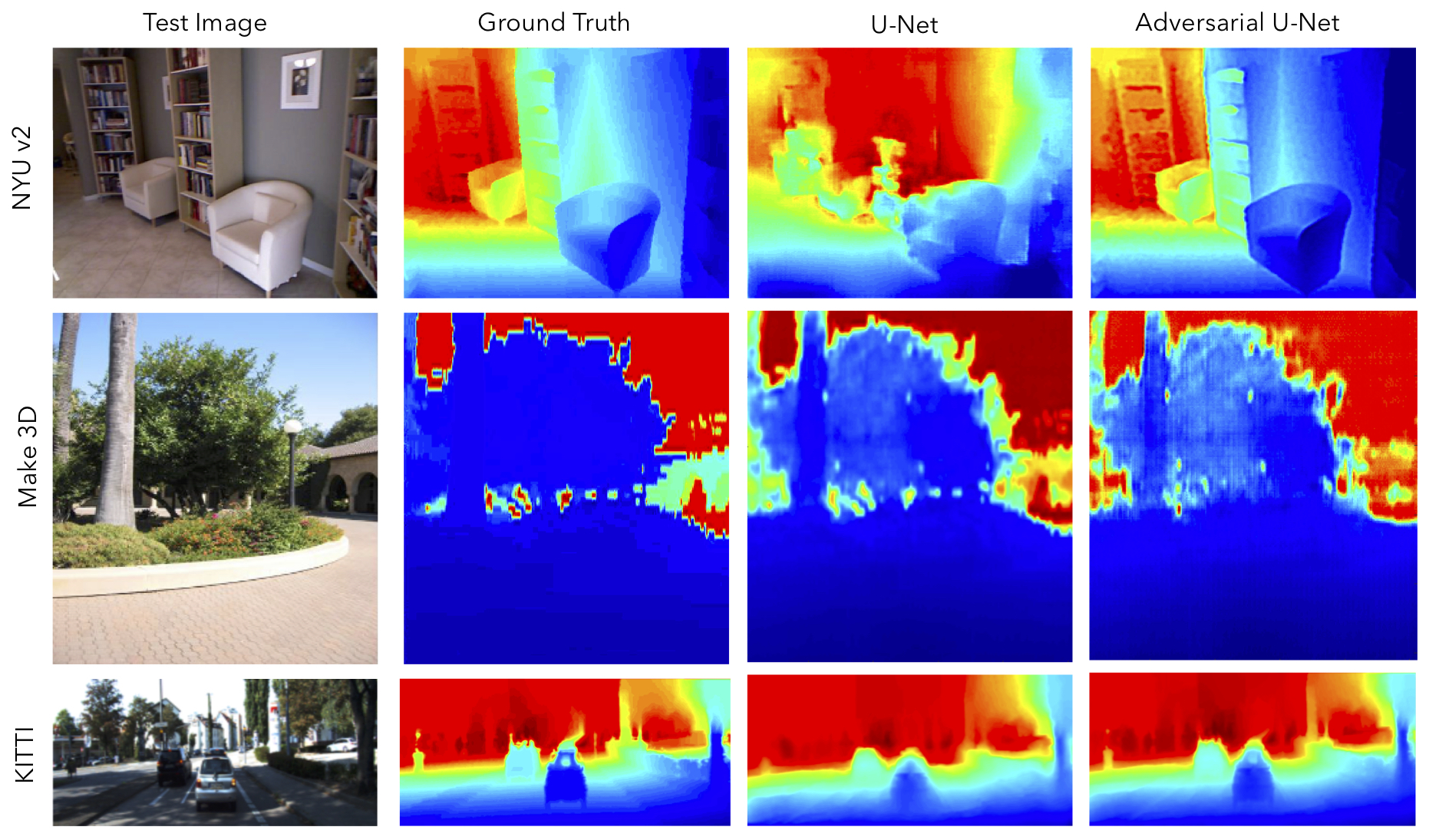}
\caption{Representative images showing estimated depth on NYUv2, Make 3D and KITTI datasets via our proposed adversarial depth estimation paradigm. We demonstrate that model trained via adversarial training can predict relievedly more accurate depths..}
\label{fig_sim}
\end{figure*}

\begin{equation}\label{eq:Tk}
  \begin{aligned}
 \phi(y_{i},y_{j};\vec\beta)= - \frac{1}{2}\sum_{k=1}^{K}\beta_{k}S_{i,j}^{k}(y_i-y_j)^2. 
  \end{aligned}
\end{equation}

The overall energy function is,

\vspace{-4mm}

\begin{equation}\label{eq:Tk}
  \begin{aligned}
 E = - \sum_{i \in \N} (y_i-h_i(\vec{\gamma}))^2  - \frac{1}{2}\sum_{(i,j) \in \Sp}\sum_{k=1}^{K}\beta_{k}S_{i,j}^{k}(y_i-y_j)^2.
  \end{aligned}
\end{equation}

\noindent During training the negative log likelihood of the probability density function is calculated from Eq. 5 and minimized with respect to the two learning parameters, $\gamma$ and $\beta$. Two regularization terms are added to the objective function to penalize heavily weighted vectors. Assuming $N$ is the number of images in the training data,

\begin{equation}\label{eq:Tk}
  \begin{aligned}
  \min_{\gamma,\beta \geq 0} {-\sum_{1}^{N}\log Pr(\vec{y}|\vec{x};\vec{\gamma},\vec{\beta}})+ \frac{\lambda_1}{2} \norm{\gamma}_2^2+ \frac{\lambda_2}{2} \norm{\beta}_2^2.
  \end{aligned}
\end{equation}

\noindent As in section 3.1, incorporating adversarial loss in this setup means treating the objective function above as the generator and jointly training the discriminator and generator using the following loss, 

$$\argminA_{G=\text{CNN-CRF}} \argmaxA_D \mathcal{L}_{\text{GAN}}(G_{\text{CNN-CRF}},D) + \lambda\mathcal{L}_{\text{L1}}(G_{\text{CNN-CRF}}).$$

\begin{table*}[!h]
\centering
\begin{tabular*}{\linewidth}{@{\extracolsep{\fill}} c||c c c|c c c}
\hline
\multirow{2}{*}{Method} & \multicolumn{3}{c|}{C1} &  \multicolumn{3}{c}{C2} \\

& rel $\downarrow$   & $log_{10} \downarrow$ & rms $\downarrow$     & rel $\downarrow$& $\text{log}_{10} \downarrow$ & rms $\downarrow$\\

\hline\hline
Make3D \cite{saxena2007learning}                & -   & -   & -            & 0.370 & 0.187 & -     \\
Liu \textit{et al.} \cite{liu2010single}           & -   & -   & -            & 0.379 & 0.148 & -     \\
DepthTransfer \cite{karsch2016depth}               & 0.355 & 0.127 & 9.20        & 0.361 & 0.148 & 15.10 \\
Liu \textit{et al.} \cite{liu2016learning}  & 0.355 & 0.137 & 9.49        & 0.338 & 0.134 & 12.60 \\
Li \textit{et al.} \cite{li2015depth}              & 0.278 & 0.092 & 7.12        & 0.279 & 0.102 & 10.27 \\
Liu \textit{et al.} \cite{liu2016learning}               & 0.287 & 0.109 & 7.36        & 0.287 & 0.122 & 14.09 \\
Roy \textit{et al.}  \cite{roy2016monocular}          & -      & -   & -            & 0.260 & 0.119 & 12.40 \\
Laina \textit{et al.} \cite{laina2016deeper}       & 0.176 & 0.072 & 4.46       & -      & -      & -      \\
LRC-Deep3D \cite{xie2016deep3d}            & 1.000  & 2.527 & 0.981         & -      & -      & -      \\
LRC \textit{et al.} \cite{godard2017unsupervised}            & 0.443 & 0.156   & 11.513    & -      & -      & -   \\
U-Net \cite{ronneberger2015u}                 & 0.428 & 0.142 & 5.127       & 0.446 & 0.164 & 6.38\\
Kuzietsov \textit{et al.} \cite{wang2015towards}      & 0.421 & 0.190 & 8.24        & -   & -      & -      \\
MS-CRF \textit{et al.}  \cite{xu2018monocular}        & 0.184 & 0.065 & 4.38        & 0.198 & -    & 8,56   \\
DORN (ResNet) \cite{fu2018deep}                    & 0.157 & 0.062 & 3.97        & 0.162 & 0.067 & 7.32  \\
\hline
CNN-CRF \cite{liu2015deep}     							& 0.314& 0.119 & 8.603     & 0.307 & 0.125 & 12.89\\
\textbf{CNN-CRF Adv.}      		& 0.287& 0.103 & 6.188     & 0.266 & 0.098 & 9.148\\
U-Net \cite{ronneberger2015u}      & 0.428 & 0.142 & 5.127       & 0.446 & 0.164 & 6.38\\
\textbf{U-Net-Adv.}              & \textbf{0.0646} & \textbf{0.0277} & \textbf{1.812} & \textbf{0.0817} & \textbf{0.0493} & \textbf{4.163}\\
\hline
\end{tabular*}
\caption{Performance on Make3D. All methods are evaluated on the test split by Make3D \cite{saxena2007learning}}
\end{table*}

\begin{table*}[!h]
\begin{center}
\begin{tabular*}{\linewidth}{@{\extracolsep{\fill}} c||c| c c c c |c c c}
\hline
Method & cap & abs rel $\downarrow$ & squared rel $\downarrow$ & rms $\downarrow$ & $\text{rms}_{\text{log}}$ $\downarrow$ & $\delta < 1.25 \uparrow$ & $\delta < 1.25^2 \uparrow$ & $\delta < 1.25^3 \uparrow$ \\
\hline\hline
Make3D \cite{saxena2007learning} 		& 0-80m & 0.280 & 3.012 & 8.734 & 0.361 & 0.601 & 0.820 & 0.926 	\\
Eigen \textit{et al.} \cite{eigen2015predicting} 		& 0-80m & 0.190 & 1.515 & 7.156 & 0.270 & 0.692 & 0.899 & 0.967		\\
Liu \textit{et al.} \cite{liu2016learning} 				& 0-80m & 0.217 & 1.841 & 6.986 & 0.289 & 0.647 & 0.882 & 0.961		\\
LRC (CS+K) \cite{godard2017unsupervised} 						& 0-50m & 0.114 & 0.898 & 4.935 & 0.206 & 0.861 & 0.949 & 0.976		\\
Kuzietsov \textit{et al.} \cite{kuznietsov2017semi} & 0-50m & 0.113 & 0.741 & 4.621 & 0.189 & 0.862 & 0.960 & 0.986		\\
DORN (VGG) \cite{fu2018deep} 			& 0-80m & 0.081 & 0.376 & 3.056 & 0.132 & 0.915 & 0.980 & 0.993		\\ 
DORN (ResNet) \cite{fu2018deep} 		& 0-80m & 0.072 & 0.307 & 2.727 & 0.120 & 0.932 & 0.984 & 0.994		\\ 
\hline
CNN-CRF \cite{liu2015deep}  								& 0-80m & 0.217 & 1.793 &  7.421 & 0.261 & 0.656 & 0.881 & 0.958	\\ 
\textbf{CNN-CRF Adv.}  			& 0-80m & 0.164 & 1.279 &  6.138 & 0.246 & 0.692 & 0.921 & 0.967	\\ 
U-Net \cite{ronneberger2015u}  									& 0-80m & 0.097 & 0.411 &  3.349 & 0.158 & 0.626 & 0.832 & 0.902	\\ 
\textbf{U-Net Adv.}  	& 0-80m & \textbf{0.061} & \textbf{0.282} &  \textbf{2.349} & \textbf{0.106} & \textbf{0.943} & \textbf{0.988} & \textbf{0.996}	\\ 
\hline	
\end{tabular*}
\end{center}
\caption{Performance on KITTI. All the methods are evaluated on the test split by Geiger et al. \cite{geiger2013vision}}
\end{table*}

\textbf{Implementation Details} We implemented our encoder-decoder network and CNN-CRF model in PyTorch and MatConvNet respectively \cite{paszke2017automatic, vedaldi2015matconvnet}, and trained on Nvidia P100 GPUs using Google Cloud. Both networks trained for 150 epochs with a base learning rate of 0.0002 with ADAM optimization in both the generator and the discriminator, followed by a linear step decay for 150 epochs. Both networks were trained from scratch, and used Xavier weight initialization and Spectral normalization in the generator and the discriminator, with the input also normalized to be between [-1, 1]. To prevent mode collapse, patches were pooled and fed to the discriminator in batches rather than individual images in each iteration. A pooling history of randomly selected 50 patch pairs was used in the discriminator. Dropout was used the bottleneck layer of our U-Net architecture, but was not added in our CNN-CRF setup.

\section{Experiments}
To demonstrate the improvements made by adversarial training relative to state-of-the-art methods, which do not use adversaries, we evaluate our methods on three standard datasets for for depth estimation: NYU Depth v2 \cite{silberman2012indoor}, Make3D \cite{saxena2007learning}, and KITTI \cite{geiger2013vision}. We also perform an ablation study for comparative analysis with and without adversarial loss.

\textbf{NYUv2.} NYUv2 is one of the largest RGB-D datasets for indoor scene reconstruction, with over 120K unique pairs of RGB and depth images acquired from 464 scenes with a Microsoft Kinect \cite{silberman2012indoor}. We worked with a 1449 aligned subset of images from NYUv2, with 795 pairs for training and 654 pairs for testing of resolution $640 \times 320$. During training, we downsampled the images to $386\times 288$, and performed random horizontal flips with random crops of size $256\times 256$. We report our scores on a pre-definend test and train split created by Eigen. \cite{eigen2015predicting}.

\textbf{Make3D.} The Make3D Range Image Dataset contains image pairs of outdoor scenes ($1704 \times 2272$) and ground truth laser depths ($55 \times 305$), with 400 pairs for training and 134 images for testing. During training, we resized the images to $400\times 300$, and performed similar random horizontal flips with random crops of $256\times 256$ crop. On this dataset, we report $C1$ and $C2$ errors (depth ranges for $0-80$m and $0-70$m respectively) using the same model.

\textbf{KITTI.} The KITTI Vision Dataset contains image pairs of outdoor scenes ($375 \times 1241$) and raw LiDaR scans for 61 scenes, ranging from "residential" to "city" scenes. We trained and tested with the 0-80 depth range for 32 scenes and 29 scenes respectively. In both splits, we preprocesed the images by resizing them to be $256 \times 256$.

\textbf{Evaluation Metrics.} Following previous works \cite{laina2016deeper,xu2018monocular,liu2010single,liu2016learning}, we considered the following performance metrics for accurate depth estimation:

\begin{enumerate}
\item Relative Error (rel): $\frac{1}{N} \sum_{y} \frac{\lvert y_{gt}-y_{est} \rvert}{y_{gt}}$
\item Average $log_{10}$ Error ($log_{10}$): $\frac{1}{N} \sum_{y} \lvert \log_{10}y_{gt}-\log_{10}y_{est} \rvert$
\item Root Mean Square Error (rms):$\sqrt{\frac{1}{N} \sum_{y} (y_{gt}-y_{est})^2}$
\item Accuracy with threshold $t$: Percentage of $y_i$ s.t. $\text{max}(\frac{y_i^*}{y_i}\frac{y_i}{y_i^*} = \delta < t(t \in [1.25, 1.25^2, 1.25^3])$, where $y_i$ is the estimated depth, $y_i^*$ is the corresponding ground truth

\end{enumerate}

In these series of comparisons, we evaluate the proposed adversarial depth estimation networks with their non-adversarial counterparts. We observed a decrease in relative error and increase in accuracy when an adversarial loss was added. In the U-Net vs. adversarial U-Net comparison, despite enforcing strong spatial consistency between the convolution and upsampling layers with long skip connections, we achieved relative errors of 0.114, 0.0646, and 0.061 on NYUv2, Make3D and KITTI respectively, which improves over the current state-of-the-art relative error by Xu \textit{et al.} \cite{xu2018monocular}. In addition, Adversarial U-Net also improves over accuracy on these three benchmarks, which suggests that the addition of adversarial training helped U-Net learn depth cues and more context-aware features that preserved structural details of the original image. Qualitatively, we can see how regular U-Net produced blurry results on the NYUv2 and KITTI datasets, however, after adding adversarial training, the network produced sharper edge details for objects in both the foreground and background. In contrast, the Adversarial CNN-CRF only marginally improved in relative error and accuracy with a threshold of 0.125 on NYUv2, with accuracy decreasing with a threshold of $0.125^3$. The low accuracy in the CNN-CRF can be attributed to how using a $\mathcal{L}_{\text{L1}}$ loss directly optimizes for the relative error rather than accuracy, which penalizes greater deviations in per-pixel predictions from the ground truth. In addition, the relatively low accuracy of adversarial CNN-CRF may be due to the small training set used to train CNN-CRF and the fact that the loss was computed on the super-pixel level rather than the entire image.

\section{Conclusions} 

In this paper, we demonstrate the effectiveness of adversarial training for monocular depth estimation from a single image across two kinds of neural network architectures: encoder-decoder style U-Net and joint CNN-CRFs. Our method approaches the depth estimation problem by incorporating an adversarial loss which captures non-local information as compared to local per-pixel losses. Unlike more complex multi-scale, deep architectures used for capturing global interactions for understanding local and non-local context, our approach is relatively more simple and robust. The global information is incorporated by a discriminator which aims to discriminate patches of an estimated depth prediction as real or fake. In our findings, we show how adversarial training can improve depth predictions as compared to state-of-the-art methods. This improvement is particularly pronounced for U-Net which performs weakly by itself, but outperforms state-of-the-art when adversarial loss is used.

\section{Acknowledgements}
This work was supported in part with funding from the NIH NIBIB Trailblazer Award (R21 EB024700).

{\small
\bibliographystyle{ieee}
\bibliography{egbib}
}

\end{document}